\documentclass{article} 
\usepackage{nips15submit_e,times}
\usepackage{hyperref}
\usepackage{url}
\usepackage{amsmath}
\usepackage{amsthm}
\usepackage{amssymb}
\usepackage{subcaption}

\usepackage{xcolor}
\usepackage[pdftex]{graphicx}

\theoremstyle{definition}

\usepackage{tikz}
\usetikzlibrary{positioning}
\usetikzlibrary{decorations.text}
\usetikzlibrary{arrows}
\usetikzlibrary{shapes}

\usepackage{pgfplots}
\usepackage{multirow}
\definecolor{mygray}{rgb}{0.29,0.3,0.3}

\title{Stratified Knowledge Bases as Interpretable Probabilistic Models (Extended Abstract)\thanks{This extended abstract describes our recent work  \cite{mln2posl,DBLP:conf/ecai/KuzelkaDS16} from the perspective of interpretable machine learning.}}

\author{Ond\v{r}ej Ku\v{z}elka
\\Cardiff University 
\\United Kingdom
\\ \scriptsize{\texttt{KuzelkaO@cardiff.ac.uk}}
\And
Jesse Davis
\\KU Leuven
\\Belgium
\\ \scriptsize{\texttt{jesse.davis@cs.kuleuven.be}}
\And
 Steven Schockaert
\\Cardiff University 
\\United Kingdom
\\ \scriptsize{\texttt{SchockaertS1@cardiff.ac.uk}}
}

\nipsfinalcopy 

\begin{document}

\maketitle

\begin{abstract}
In this paper, we advocate the use of stratified logical theories for representing probabilistic models. We argue that such encodings can be more interpretable than those obtained in existing frameworks such as Markov logic networks. Among others, this allows for the use of domain experts to improve learned models by directly removing, adding, or modifying logical formulas.
\end{abstract}

\section{Introduction}
The use of symbolic representations for encoding generative models is appealing, as they can be more transparent than black-box models such as neural networks. One notable example of such a symbolic framework is Markov logic networks (MLNs \cite{MLN}), which use weighted first-order formulas to encode probabilistic models. These logical formulas can give us some important insights into what regularities have been identified in a given dataset. Unfortunately, the weights that are used in MLNs can be hard to interpret, as their intuitive meaning depends on the interplay with the weights of other formulas \cite{Thimm:2014b,mln2posl}. This means that we cannot readily use MLNs to generate explanations for predictions (as the formulas influence cannot be interpreted in isolation), and this makes it difficult for domain experts to directly verify or improve learned MLNs.  

Density estimation trees \cite{ram2011density} encode probabilistic models in a symbolic way by explicitly defining sets of possible worlds (i.e.\ those that satisfy all the literals associated with a given branch of the tree) that have the same probability (i.e.\ the value that is associated with the leaf of that branch). Since each branch can be interpreted separately from the rest of the tree, a domain expert can readily verify whether a learned model is reasonable. However, the use of a tree structure means that relationships between variables are not encoded explicitly. For example, it may be that branches where $a$ is true and $b$ is false always have a very low probability, but this can only be verified by inspecting a large number of branches. 

The approach we propose starts by converting a density estimation tree into a stratified logical theory, and then uses different forms of pruning to obtain an (often significantly) more compact theory, in which relationships between variables such as ``if $a$ then typically $\neg b$'' are modelled explicitly \cite{DBLP:conf/ecai/KuzelkaDS16}. This method is explained in more detail in the next section.


\section{Logical encoding of density estimation trees}

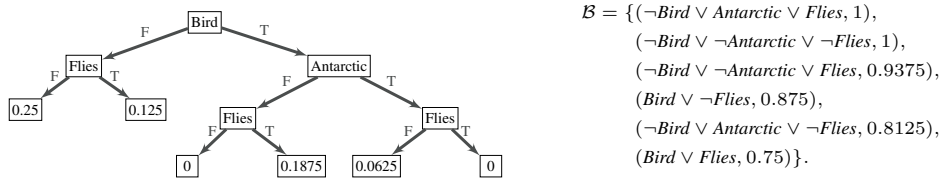
\begin{figure}[t]
\begin{subfigure}{0.48\textwidth}
\centering
\resizebox{1\textwidth}{!}{
\tikzset{
main node/.style={rectangle,fill=white!30,draw,minimum size=0.5cm,inner sep=2pt},
node 1/.style={circle,fill=red!20,draw,minimum size=0.5cm,inner sep=0pt},
node 2/.style={circle,fill=green!20,draw,minimum size=0.5cm,inner sep=0pt},
node 3/.style={circle,fill=yellow!20,draw,minimum size=0.5cm,inner sep=0pt},
node 4/.style={circle,fill=black!20,draw,minimum size=0.5cm,inner sep=0pt},
}
\tikzset{edge/.style = {->,> = latex'}}
\begin{tikzpicture}

\node[main node] (1) {Bird};
\node[main node] (2) [below left = 0.5cm and 2cm of 1] {Flies};
\node[main node] (3) [below right = 0.5 cm and 2cm of 1] {Antarctic};
\node[main node] (4) [below left = 0.5 cm and 0.55cm of 2] {0.25};
\node[main node] (5) [below right = 0.5 cm and 0.55cm of 2] {0.125};
\node[main node] (6) [below left = 0.7 cm and 1.2cm of 3] {Flies};
\node[main node] (7) [below right = 0.7 cm and 1.2cm of 3] {Flies};

\node[main node] (12) [below left = 0.6 cm and 0.5cm of 6] {0};
\node[main node] (13) [below right = 0.6 cm and 0.5cm of 6] {0.1875};
\node[main node] (14) [below left = 0.6 cm and 0.5cm of 7] {0.0625};
\node[main node] (15) [below right = 0.6 cm and 0.5cm of 7] {0};

\draw[edge, line width = 2pt, color = mygray] (1) to  node[above] {F} (2);
\draw[edge, line width = 2pt, color = mygray] (1) to node[above] {T} (3);
\draw[edge, line width = 2pt, color = mygray] (2) to node[above] {F} (4);
\draw[edge, line width = 2pt, color = mygray] (2) to node[above] {T} (5);
\draw[edge, line width = 2pt, color = mygray] (3) to node[above] {F} (6);
\draw[edge, line width = 2pt, color = mygray] (3) to node[above] {T} (7);

\draw[edge, line width = 2pt, color = mygray] (6) to node[above] {F} (12);
\draw[edge, line width = 2pt, color = mygray] (6) to node[above] {T} (13);
\draw[edge, line width = 2pt, color = mygray] (7) to node[above] {F} (14);
\draw[edge, line width = 2pt, color = mygray] (7) to node[above] {T} (15);

\end{tikzpicture}}
\end{subfigure}
\begin{subfigure}{.48\textwidth}
{\scriptsize
\begin{align*}
 \mathcal{B} =  \{&(\neg \textit{Bird} \vee \textit{Antarctic} \vee \textit{Flies}, 1),\\
& (\neg \textit{Bird} \vee \neg \textit{Antarctic} \vee \neg \textit{Flies}, 1), \\
& (\neg \textit{Bird} \vee \neg \textit{Antarctic} \vee \textit{Flies}, 0.9375),\\
& (\textit{Bird} \vee \neg \textit{Flies}, 0.875), \\
& (\neg \textit{Bird} \vee \textit{Antarctic} \vee \neg \textit{Flies}, 0.8125),\\
& (\textit{Bird} \vee  \textit{Flies}, 0.75)\}.
\end{align*}}
\end{subfigure}
\caption{A density estimation tree (left) and its corresponding stratified knowledge base (right).\label{fig:tree1}}
\end{figure}

We can straightforwardly transform a density estimation tree into a weighted logical theory, containing one formula for each branch of the tree. This transformation is illustrated in Figure \ref{fig:tree1}. The density estimation tree on the left-hand side encodes, among others, that every world in which the atoms \textit{Bird} and \textit{Flies} are false has a probability of 0.25. The representation on the right-hand side of Figure \ref{fig:tree1} encodes the same knowledge using possibilistic logic \cite{DLP}. Specifically, each branch of the density tree can be seen as a conjunction of literals; e.g.\ the left-most branch encodes the conjunction $\neg \textit{Bird} \wedge \neg \textit{Flies}$. 
For the conjunction $\alpha$ and the probability $p$ associated with each branch in the tree, the possibilistic logic theory on the right-hand side contains the weighted formula $(\neg \alpha, 1-p)$. 
For example, the left-most branch has $\alpha=\neg \textit{Bird} \wedge \neg \textit{Flies}$ and $p=0.25$, and thus the weighted formula $(\textit{Bird} \vee  \textit{Flies}, 0.75)$ appears in the theory.
In general, a possibilistic logic theory $\{(\alpha_1,\lambda_1),...,(\alpha_n,\lambda_n)\}$, with each $\alpha_i$ a formula and each $\lambda_i$ a certainty weight ($0<\lambda_i\leq 1$), encodes a distribution $\pi$ over possible worlds, defined for a possible world $\omega$ as $\pi(\omega)=1-\lambda_i$, with $\lambda_i$ the highest weight of a formula $\alpha_i$ that is violated by $\omega$, i.e.\ $\pi(\omega)=1-\max\{\lambda_i \,|\, \omega\not\models \alpha_i\}$. For a possible world $\omega$, $\pi(\omega)$ is called the possibility degree of $\omega$. Thus each weighted formula $(\alpha_i,\lambda_i)$ can be seen as encoding an upper bound of $1-\lambda_i$ on the value of $\pi(\omega)$ for all worlds $\omega$ that violate $\alpha_i$. 

Possibility theory does not impose any particular meaning on the possibility degrees. Note in particular that $\pi$ is typically not a probability distribution. In fact, in many applications, possibility degrees are interpreted in a purely ordinal fashion. A possibilistic logic theory is then simply seen as a ranking of formulas, i.e.\ a stratified knowledge base (SKB), which induces a ranking on possible worlds. However, when constructing possibilistic logic theories from density estimation trees, as in the example from Figure \ref{fig:tree1}, the associated distribution $\pi$ is actually a probability distribution. We will refer to possibilistic logic theories that induce a probability distribution as stratified probabilistic knowledge bases (SPKBs). We will use the terms SPKB and SKB to highlight whether we talk about possibilistic logic theories that induce a probability distribution or merely a ranking of possible worlds. As we will see in the next section, among others, SKBs can be used for maximum a posteriori (MAP) inference, while SPKBs can additionally be used for computing marginal probabilities.



\paragraph{Pruning}
An important advantage of possibilistic logic representations, over density estimation trees, is that we can use logical inference to derive representations which are more compact, and which encode the regularities underlying the data more explicitly. In particular, let us write $\mathcal{B}_{\lambda}$ for the set of classical formulas in a possibilistic logic theory $\mathcal{B}$ whose weight is at least $\lambda$. If a formula $\alpha$ can be derived from $\mathcal{B}_{\lambda}$ then $\mathcal{B}$ and $\mathcal{B}\cup\{(\alpha,\lambda\}$ are equivalent, in the sense that they induce the same possibility distribution $\pi$. Consequently, a standard SAT solver can be used to remove weighted formulas that are redundant and to simplify the remaining formulas (i.e.\ remove redundant literals). 
For example, the possibilistic logic theory from Figure \ref{fig:tree1} can be equivalently written as:
\begin{align*}
\mathcal{B}' =  \{&(\neg \textit{Bird} \vee \textit{Antarctic} \vee \textit{Flies}, 1),(\neg\textit{Bird} \vee \neg\textit{Antarctic} \vee \neg \textit{Flies}, 1), \\
&(\neg\textit{Bird} \vee \neg \textit{Antarctic}, 0.9375), (\neg \textit{Flies} \vee \textit{Bird}, 0.875),(\neg \textit{Bird}, 0.8125)\}
\end{align*}
where e.g.\ $(\neg \textit{Bird} \vee \neg \textit{Antarctic} \vee \textit{Flies}, 0.9375)$ was rewritten as $(\neg\textit{Bird} \vee \neg \textit{Antarctic}, 0.9375)$, because we also have the formula $(\neg \textit{Bird} \vee \neg\textit{Antarctic} \vee \neg\textit{Flies}, 1)$. In particular, because $\neg \textit{Bird} \vee \neg\textit{Antarctic} \vee \neg\textit{Flies}$ and $\neg\textit{Bird} \vee \neg \textit{Antarctic} \vee \textit{Flies}$ imply $\neg \textit{Bird} \vee \neg \textit{Antarctic}$, this means that $(\neg \textit{Bird} \vee \neg \textit{Antarctic}, 0.9375)$ can be added to $\mathcal{B}$, after which the original formula $(\neg \textit{Bird} \vee \neg \textit{Antarctic} \vee \textit{Flies}, 0.9375)$ becomes redundant. While the change between $\mathcal{B}$ and $\mathcal{B}'$ is perhaps not drastic, we have shown in \cite{DBLP:conf/ecai/KuzelkaDS16} that (in extreme cases) this method can lead to theories which are exponentially more compact than the initial density estimation tree. Moreover, apart from this exact form of logical simplification, there are also several approximate ways in which we can further simplify the possibilistic logic theory, such that the possibility distribution induced by the resulting theory stays close to the distribution induced by the initial tree. For example, one effective strategy consists in merging the levels with the highest certainty degrees, whose only effect is that the probabilities of the least probable worlds can no longer be differentiated, which often does not matter much in applications. Experimental results in \cite{DBLP:conf/ecai/KuzelkaDS16} show that thus reducing the possibilistic logic theories to 10\% of their initial size usually does not lead to significantly lower predictive accuracies for MAP-inference.

\paragraph{Interaction with domain experts}
Another important advantage of possibilistic logic representations is that they can be easily modified by human users. It is then often convenient to convert the formulas into implications\footnote{When converting a clause $C$ into an implication, we select the antecedent $\alpha$ and consequent $\beta$ (where $C = \neg \alpha \vee \beta$) so that $\beta$ would be true in all most probable worlds where $\alpha$ is true. In other words, using terminology of Section \ref{sec:inference}, we select $\alpha$ and $\beta$ so that $\alpha$ would MAP-entail $\beta$, i.e.\ so that $\alpha$ would be sufficient evidence to derive $\beta$ using MAP-inference. In cases when this is not possible, we leave $C$ in the clausal form.}, e.g.:
\begin{align*}
\mathcal{B}'' =  \{&(\textit{Bird} \wedge \neg \textit{Antarctic} \rightarrow \textit{Flies}, 1),(\textit{Bird} \wedge \textit{Antarctic} \rightarrow \neg \textit{Flies}, 1), \\
&(\textit{Bird} \rightarrow \neg \textit{Antarctic}, 0.9375), (\textit{Flies} \rightarrow \textit{Bird}, 0.875),(\neg \textit{Bird}, 0.8125)\}
\end{align*}
We may have an expert telling us that we should not be deriving with full certainty that Antarctic birds do not fly, as e.g.\ there are albatrosses living in Antarctica. It is easy to see that we can achieve this by simply removing the rule $(\textit{Bird} \wedge \textit{Antarctic} \rightarrow \neg \textit{Flies}, 1)$, although some bookkeeping is needed to check whether any earlier pruning steps need to be revised. Similarly, we may have an expert telling us that the relative certainty of two formulas should be reversed. For example, if the expert tells us that it is more common to see an Antarctic bird than it is to see a flying creature that is not a bird, this tells us that the formula $\textit{Flies} \rightarrow \textit{Bird}$ should appear with a higher certainty degree than the formula $\textit{Bird} \rightarrow \neg \textit{Antarctic}$. In ordinal settings this is straightforward, but in SPKBs it requires us to relearn suitable certainty degrees, subject to the ranking constraint provided by the expert. In \cite{DBLP:conf/ecai/KuzelkaDS16} we show how such weights can be found using geometric programming.

\section{Inference}\label{sec:inference}

\paragraph{Inference with SKBs}
From an SKB we can only obtain an ordering on possible worlds. However, if the SKB orders the possible worlds according to their probability, this is enough to evaluate MAP queries of the form ``what formulas are true in the most probable models of $\alpha$''. In fact, it turns out that MAP inference corresponds to a standard technique for inconsistency-tolerant reasoning in possibilistic logic. In particular, to verify whether a formula $\beta$ is true in all the highest-ranked models of $\alpha$, w.r.t.\ a possibilistic logic theory $\mathcal{B}$, we can proceed as follows. First we add $(\alpha,1)$ to $\mathcal{B}$. Then we find the minimal weight $w \in [0,1]$ such that the set of formulas which appear with a weight that is strictly higher than $w$ forms a consistent set $\mathcal{B}_{\alpha}$ of classical formulas. It can be shown that a formula $\beta$ holds in all the highest-ranked models of $\alpha$ iff it can be derived from $\mathcal{B}_{\alpha}$. If that is the case, we say that $\beta$ is MAP-entailed by $\alpha$ and write $\alpha \vdash_{\mathcal{B}} \beta$. Note that we can check whether $\alpha \vdash_\mathcal{B} \beta$ using a logarithmic number of calls to a SAT solver (in the number of different certainty degrees appearing in $\mathcal{B}$). We now illustrate the procedure using the following toy example:
\begin{align*}
\mathcal{B} = \{ (\textit{Gardener} \rightarrow \neg \textit{HayFever}, 0.9), (\textit{Coughs} \rightarrow \textit{HayFever}, 0.8) \}.
\end{align*}
It holds that $\textit{HayFever} \vdash_\mathcal{B} \neg \textit{Gardener}$, which follows from the fact that $\{\textit{HayFever},\textit{Gardener} \rightarrow \neg \textit{HayFever},\textit{Coughs} \rightarrow \textit{HayFever}\}$ is consistent (i.e.\ the minimal weight $w$ that ensures consistency is 0), and $\textit{HayFever}$ and $\textit{Gardener} \rightarrow \neg \textit{HayFever}$ entail $\neg \textit{Gardener}$. We also have $\textit{Gardener} \wedge \textit{Coughs} \vdash_\mathcal{B} \neg \textit{HayFever}$, which can be seen as follows. First, if we add $(\textit{Gardener} \wedge \textit{Coughs},1)$ to $\mathcal{B}$ then it becomes inconsistent as a propositional logic theory because $\textit{Gardener}$ implies $\neg \textit{HayFever}$ but $\textit{Coughs}$ implies $\textit{HayFever}$, which is a contradiction. Following the aforementioned procedure, we find $w=0.8$ and accordingly remove the rule $(\textit{Coughs}\rightarrow \textit{HayFever}, 0.8)$ and obtain  $\mathcal{B}_{\alpha} = \{\textit{Gardener} \rightarrow \neg \textit{HayFever}, \textit{Gardener} \wedge \textit{Coughs}\}$, from which we can indeed derive $\neg \textit{HayFever}$.

Beyond MAP inference, we can also derive e.g.\ what is true in the top 20\% most probable models of $\alpha$ using a model counter. Let $w_1,...,w_k=1$ be the certainty weights appearing in $\mathcal{B} \cup \{(\alpha,1)\}$ that are strictly higher than the cut-off value $w$ from the previous procedure. Let $A_i$ be the set of formulas appearing with weight $w_i$. A formula can be derived from $A_l \cup ... \cup A_k$ iff it is true in the top $\theta\%$ most probable models of $\alpha$ with $\theta = \frac{M(A_l\cup...\cup A_k)}{M(\alpha)}$, where for a set of formulas $E$ we write $M(E)$ for its number of models. This type of query is very natural, yielding conclusions which are similar in spirit to statements such as ``$90\%$ of adult men weigh between 70 and 100kg'', ``80\% adult men weigh between 75 and 95 kg'', etc. Indeed, as we can consider nested intervals of increasingly typical weights, we can similarly characterize formulas that specify increasingly typical possible worlds.

\paragraph{Inference with SPKBs}
SPKBs support a number of additional query types. For example, for a probability threshold $t$, the worlds whose probability is greater than $t$ are those which satisfy all formulas that appear in the considered SPKB $\mathcal{B}$ with a weight $w \geq 1-t$.  Moreover, based on this insight, we can use a SPKB $\mathcal{B}$ to compute the marginal probability $P(\alpha)$ of any propositional formula. In particular, we can show that \cite{DBLP:conf/ecai/KuzelkaDS16}:
\begin{align*}
P(\alpha) = \sum_{i = 1}^{k} (1-w_i) \cdot (M(B_{i+1)})- M(B_i))
\end{align*}
where $w_1,...,w_k$ are the certainty degrees appearing in $\mathcal{B}$ (sorted in increasing order); for $i\leq k$, $B_{i}$ is the set of formulas that appear in $\mathcal{B} \cup \{(\alpha,1)\}$ with weight at least $w_i$, while $B_{k+1} = \{\alpha\}$; $M(.)$ again represents the model count operator.

\section{Conclusions and future work}

Stratified (probabilistic) knowledge bases can encode the information captured by density estimation trees in a logical form. This leads to several important advantages for interpretable machine learning. First, SKBs can be used to generate logical justifications (proofs) to support the predictions that are made from them. Moreover the explicit symbolic nature of SKBs allows users, with a basic understanding of propositional logic, to relatively easily modify SKBs. These properties make SKBs stand out not only when compared to black-box models such as neural networks, but even when compared to other symbolic methods such as Markov logic networks.

Several interesting directions for future work remain. For example, in \cite{mln2posl} we have shown that, for any probability distribution, it is possible to construct a compact SKB which exactly captures MAP-inference with evidence sets of bounded size. The techniques presented in \cite{mln2posl} can naturally be applied for summarizing what can be derived from density estimation trees when only small evidence sets are considered. In practice, this can lead to even more compact and more easily interpretable SKBs, providing a more local view of the corresponding distributions. 
Another interesting direction would be to lift our approach into the first-order setting, similar to how we lifted the method in \cite{mln2posl} to encode MLNs. Finally, note that while we have only considered binary properties in this paper, density estimation trees can also handle continuous variables, which means that adding support for continuous variables in our setting is relatively straightforward.




\section*{Acknowledgment}
This work has been supported by a grant from the Leverhulme Trust (RPG-2014-164). Jesse Davis is partially supported by the KU Leuven Research Fund (C22/15/015), and FWO-Vlaanderen (G.0356.12, SBO-150033).

\bibliography{ecai,main}
\bibliographystyle{abbrv}

\end{document}